\pdfoutput=1
\documentclass[11pt]{article}

\usepackage[]{emnlp2023}

\usepackage{times}
\usepackage{latexsym}
\usepackage{booktabs}
\usepackage{graphicx}
\usepackage{tabularx}
\usepackage{longtable}
\usepackage{array}
\usepackage{enumitem}

\usepackage[T1]{fontenc}
\usepackage[utf8]{inputenc}

\usepackage{microtype}

\usepackage{inconsolata}

\title{A Few Hypocrites: Few-Shot Learning and Subtype Definitions for Detecting Hypocrisy Accusations in Online Climate Change Debates}

\author{Paulina Garcia Corral\textsuperscript{1}, Avishai Green\textsuperscript{2}, Hendrik Meyer\textsuperscript{3}\textsuperscript{*} \\
\textbf{Anke Stoll}\textsuperscript{4}, \textbf{Xiaoyue Yan}\textsuperscript{5}, \textbf{Myrthe Reuver}\textsuperscript{6}\\
\textsuperscript{1} Data Science Lab, Hertie School
\textsuperscript{2} Hebrew University of Jerusalem
\textsuperscript{3} University of Hamburg\\
\textsuperscript{4} Ilmenau University of Technology
\textsuperscript{5} University of Zurich
\textsuperscript{6} Vrije Universiteit Amsterdam\\
 \textsuperscript{1} \texttt{corral@hertie-school.org}, \textsuperscript{2} \texttt{avishai.green@mail.huji.ac.il} \\ \textsuperscript{3}\texttt{hendrik.meyer-1@uni-hamburg.de}\\
  \small{* The first three authors contributed equally, and are shared first authors.}\\}

\begin{document}
\maketitle
\begin{abstract}
The climate crisis is a salient issue in online discussions, and hypocrisy accusations are a central rhetorical element in these debates. However, for large-scale text analysis, \textit{hypocrisy accusation detection} is an understudied tool,  most often defined as a smaller subtask of fallacious argument detection. In this paper, we define hypocrisy accusation detection as an independent task in NLP, and identify different relevant subtypes of hypocrisy accusations. Our Climate Hypocrisy Accusation Corpus (CHAC) consists of 420 Reddit climate debate comments, expert-annotated into two different types of hypocrisy accusations: personal versus political hypocrisy. We evaluate few-shot in-context learning with 6 shots and 3 instruction-tuned Large Language Models (LLMs) for detecting hypocrisy accusations in this dataset. Results indicate that the GPT-4o and Llama-3 models in particular show promise in detecting hypocrisy accusations (F1 reaching 0.68, while previous work shows F1 of 0.44). However, context matters for a complex semantic concept such as hypocrisy accusations, and we find models struggle especially at identifying political hypocrisy accusations compared to personal moral hypocrisy. Our study contributes new insights in hypocrisy detection and climate change discourse, and is a stepping stone for large-scale analysis of hypocrisy accusation in online climate debates.
%by defining types of hypocrisy accusations for computational analysis, presenting a novel dataset, and evaluating LLM's performance in this context.
\end{abstract}

\section{Introduction}\label{sec:introduction}

Perhaps no accusation is more commonly lobbied in political discourse as that of hypocrisy \citep{Collins2018, Thompson2004}. Allegations of hypocrisy, defined by the Oxford English Dictionary as the assumption ``of a false appearance of virtue or goodness'' \cite{OED}
and understood in practice as an incongruity between behavior and publicly expressed beliefs \cite{Furia2009}, are so ubiquitous as to lead Hannah Arendt to describe politics itself as a ``never ending fight to ferret out hypocrites'' \cite{Arendt2006}. Making such an accusation of one's rival is effective, as hypocrisy is widely and deeply loathed, with research showing that perceived hypocrisy negatively affects voters’ opinions of politicians above and beyond other underlying scandals \citep{Bhatti2013, Laurent2014, Grover2015}. It  may also seem, in a polarized political landscape, like the only rhetorical tool available. When political opponents lack shared standards, moral persuasion becomes near-impossible, and what is left is undermining one’s opponent with “the revelation that… [he or she] is not living up to his own professed ideal” \citep{Shklar1984} . 
While discourse on political hypocrisy dates centuries back, it has only intensified online, buoyed by social media’s amplification of polarization \citep{Allcott2020} and valorization of authenticity \citep{Hallinan2021}. 

An arena of online discourse in which the hypocrisy charge is especially pertinent is that of debates around climate change, where hypocrisy accusations have been shown to be central to increasingly polarized, cross-ideological online interactions \citep{BrueggemannMeyer2023}. An example of such an online hypocrisy accusation is a Reddit commenter describing COP26, an international political climate conference in 2020, as \textit{``the biggest hypocrisy in the world''} [because the politicians are] \textit{``arriving in their private jets.''} Hypocrisy accusations have been found to drive polarization in online climate debates \citep{Falkenberg2022}, facilitating not cross-camp deliberation but rather segregation into opposing ideological camps \citep{Meyer2023} and affective polarization \citep{9381419}. Being able to analyze (online) debates about climate policy, focusing on hypocrisy accusations can help us better measure the use of hypocrisy in such debates. %and to understand how and to what effect it is used as a rhetorical tool.

Detecting accusations of hypocrisy is understudied in computational text analysis and Natural Language Processing (NLP). Usually, hypocrisy is only included as part of broader logical fallacy detection tasks, as in \citet{Habernal2018}. However, hypocrisy accusations are a distinct phenomenon: Linguistically, they often contain contrastive conjunctions between clauses to highlight inconsistencies. Semantically, their identification relies on context to understand contradictions. %that signal accusations. 

Earlier work (see Section \ref{sec:background}) reports good performance on detecting \textit{logical fallacies} in general, but it is noticeable in the literature that hypocrisy accusation are a difficult and sometimes neglected phenomenon in state-of-the-art NLP. \citet{alhindi-etal-2022-multitask} report relatively low performance of 0.43 in macro-F1 for detecting \textit{whataboutism}, a hypocrisy-related fallacy – and only 0.21 in macro-F1 for all fallacies in climate debates. Recently, \citet{piskorski-etal-2023-multilingual} report a performance of RoBERTa-XLM multilingual detection of the \textit{whataboutism} concept with an F1 of 0.06. However, this concept consists of only 0.05\% of their fallacy dataset.

%and an F1 score of only 0.29 for hypocrisy accusations with the T5 Large Language Model (LLM).

Due to a lack of attention to the specific \textit{hypocrisy accusation} construct, there is a scarcity of data annotated for hypocrisy detection as well as little research of nuances in this construct, such as different types of hypocrisy accusations. Yet hypocrisy allegations are varied, and we believe an account of different kinds will afford a better understanding of the online debate and the performance of models detecting such accusations.
%Political discourse is a low-resource domain in NLP, and hypocrisy accusations have very recently been introduced as a social construct to analyze in political communication. To conduct our study, 
%Therefore, we present the  Climate Hypocrisy Accusation Corpus (CHAC), with different types of hyporcisy accusations based on political science theory: personal moral hypocrisy and political hypocrisy. Leveraging LLM's in-context learning capabilities, we evaluate how well LLMs identify these different types of accusations of hypocrisy in climate debates. These accusations tend to include the use of irony, sarcasm and other rhetorical devices to highlight the moral inconsistency of actors, and contextual knowledge is key to understanding the contrast that is being highlighted. We analyze the potential of LLMs as hypocrisy accusation detectors, and discuss the various hypocrisy constructions that we find in our data.

Our contributions to the literature are:  \\ (1) We, unlike previous work, analyze \textbf{hypocrisy accusation detection as an individual and nuanced task} particularly relevant for analyzing online climate discourse; \\
(2) We are also, to our knowledge, the first to define \textbf{different types of hypocrisy} accusation for computational analysis, where we differentiate personal moral hypocrisy from political hypocrisy; \\
(3) Additionally, we release a \textbf{dataset}: the  Climate Hypocrisy Accusation Corpus (CHAC), with 420 comments annotated by social scientists for the different types of hypocrisy accusations; \\ 
(4) We \textbf{analyze the potential of Large Language Models} as hypocrisy accusation detectors and evaluate LLMs with the various hypocrisy constructions we find in our data. We note where models struggle with the task, as a building block for future hypocrisy accusation analyses.

\vspace{0.2cm}

This paper is as organized as follows: Section \ref{sec:background} describes previous research on measuring social science constructs using LLMs, including  hypocrisy detection. Section \ref{sec:ocach-creation} introduces our dataset: the different types of hypocrisy, our annotation process, and corpus statistics. We describe our experiments to evaluate the capabilities of different instruction-tuned models on this dataset in Section \ref{sec:experimental-setup}, and present results of these experiments in Section \ref{sec:results}. We then reflect on our results and the complexity of our task (Section \ref{sec:discussion}), and conclude our work in Section \ref{sec:conclusions}.

\section{Background}\label{sec:background}

Large Language Models (LLMs) are models trained on predicting sequences of text. These models are trained in human preferences and instructions, and can perform in-context learning: with a natural language prompt or instruction, these models are able to do tasks such as classifying text examples on a new construct \cite{brown2020language}.

\subsection{Promise and Limits of LLMs for Social Science Construct Detection}

LLMs show great promise for detecting complex social science constructs in text. Recent comparative analyses highlight their exceptional performance and adaptability across numerous NLP tasks, including but not limited to sentiment analysis, offensive language detection, intent recognition, fake news classification, stance detection, and document classification \cite{10380590}. \citet{alizadeh2024opensource} compared the performance of open-source LLMs in text classification tasks typical for political science research, employing both zero-shot and fine-tuned LLMs for tasks including stance, topic, and relevance classification on news articles and tweets. They concluded that fine-tuning enhances the performance of open-source LLMs, making it preferable to few-shot training with a relatively modest quantity of annotated text. \citet{Trnberg2023ChatGPT4OE} explored the utility of GPT models for annotating political Twitter messages and found that ChatGPT-4o achieved higher accuracy, reliability, and either equal or lower bias compared to human classifiers. It also excelled in annotations requiring contextual reasoning and inference of authorial intent. 

However, these models show some potential limitations.
\citet{plaza2023respectful} have found LLMs using in-context learning to outperform other NLP models in detecting complex social constructs such as sexist comments and misogynist hate, though they do report prompt brittleness – resulting in less stability of construct detection over slightly different prompt formulations. Additionally, humans remain better than LLMs at editing and improving difficult examples for model training, e.g. in sexist language detection \cite{sen2023people}. This indicates that LLMs are promising for the annotation and detection of social science construct in text, but also show some limitations in subtle understanding of the underlying construct.

\subsection{Fallacies and Hypocrisy Accusations}
Hypocrisy accusations are a subtle construct, but are not often the main topic in research using state-of-the-art NLP. \citet{Habernal2018} identify fallacies as unfair arguments in debates, `deceptions in disguise' whose conceptualization goes back to Aristotle
\cite{Aristotle1909}. Their work locates hypocrisy accusations as a subtype of \textit{ad hominem} fallacy, and find a convolutional neutral network is able to detect \textit{ad hominem} fallacies with 0.81 accuracy.  \citet{sahai2021breaking} mention hypocrisy accusations as a common fallacy in online debate, but do not include it as one of the eight fallacy types they detect with neural models with 0.76 accuracy.

Hypocrisy has also been part of other types of tasks. \citet{piskorski-etal-2023-multilingual} introduce a multilingual dataset with an annotated hypocrisy accusation concept as part of a `persuasion techniques task'. They also introduce an XLM-RoBERTa model as a baseline. One of the topics in their dataset is the climate change debate. Their appendix reports a performance of the \textit{whataboutism} concept of 0.25\% precision, with extremely low recall (0.034\%) leading to an F1 of 0.06. However, this concept is only 0.05\% of the entire dataset.

More recently, fallacy detection has also been explored with LLMs. The Logical Fallacy Understanding Dataset (LFUD) \cite{li2024reason} was created to evaluate LLMs' capability of logical fallacy understanding. The authors show how this dataset can be fine-tuned to obtain significantly enhanced performance on logical reasoning. On a limited set of logical fallacies (\textit{Against the Person, Appeal to Authority, Appeal to Popularity, Appeal to Emotion, Hasty Generalization, Questionable Cause, and Red Herring}), GPT-4o achieves an accuracy of 0.79, and when used in cases that exclude invalid or unidentified instances, an accuracy of 0.90 \cite{lim2024evaluation}. Additionally, \citet{Valdovinos_2023} created a real-time fallacy detection for events such as presidential debates online, which integrates audio transcription models with four fallacy classification models.  However, there is no category of hypocrisy accusations in these previous LLM works, nor do they focus on climate debates.

\citet{Alhindi2023} apply LLMs for detecting fallacies in online climate debates, using five existing fallacy datasets as well a new dataset constructed specifically for the task. 
%In their annotation scheme, a hypocrisy accusation is only one sub-category of fallacy. %In one of their datasets, the Propaganda dataset, ``whataboutism'' is defined as ``a discrediting technique that accuses others of hypocrisy''. 
Their experimental set-up consists of fine-tuning different sizes of the T5 LLM model on five fallacy datasets with different fallacies (and different topics of debate, from COVID-19 to the climate) before using in-context learning with prompts for detecting fallacies in the target dataset. Using this training scheme, they are able to detect \textit{whataboutism} in one dataset with 0.44 accuracy. %and adding a definition of this subclass leads to a small reduction of performance to 0.43.
Their climate dataset contains no hypocrisy class, and sees an average performance of 0.21 in macro-F1 over nine other fallacies. They acknowledge that context is essential for understanding both the climate debate and \textit{whataboutism}.

Thus, hypocrisy accusations have so far not received sufficient attention in either dataset creation or model development. Yet hypocrisy accusation detection is a complex task in its own right, with both semantic and logical context needed for success, and complexity added by several social language factors such as irony and sarcasm. Additionally, earlier results lack a careful evaluation of which different types of hypocrisy accusations can be detected. In our dataset and experiments, we intend to fill this gap by presenting a specialized climate hypocrisy accusation dataset, and an analysis of the performance of currently popular LLMs in detecting hypocrisy accusations.

\section{Data}\label{sec:ocach-creation}
We present a dataset based on the English-language \textit{Reddit European Sustainability Initiatives} corpus released by \citet{Reuver}. This corpus consists of 2,073 sustainability discussions from between 2017 and 2022 on the Reddit.com sub-communities \cite{proferes2021studying} called \textit{europe}, \textit{europeanunion}, and \textit{europes}, with 46,285 comments. Nearly half (922) of these discussions have at least one comment.

We focus on the comments in this dataset, as they constitute active discourse between users on the identified discussion topics, which are relevant for hypocrisy accusations. This means that our unit of analysis is a comment, which can contain a single or multiple sentences. 

\subsection{Data Sample and Annotation Process}\label{ssec:data}
Our sample selection involves two main strategies. We divided the data into two groups: 1) instances were hypocrisy was explicitly mentioned, by using the regex pattern \verb|hypocr*|, and 2) the remaining data. Subsequently, we randomly selected 300 samples from each group, and consolidated them into a single dataset with 600 samples. The sampling strategy was done due to the relative rarity of the explicit hypocrisy mentions.

The six expert annotators were the authors of this study, all experts in political science, environmental communication science, or (computational) linguistics, which allowed for thorough, high-quality annotations. All participated in a test round to test the annotation scheme and make comments and adjustments. After flagged issues were solved, we then proceeded to annotate the final dataset. 

Each expert annotated half of the final dataset, which yielded 3 annotations per sample. This sample size also aligns with existing literature, which suggests that a dataset of this magnitude is generally sufficient for few-shot learning tasks, offering a balanced measure of model performance with respect to human annotation capabilities. 
%This approach aims to oversample explicit expressions, given how relatively infrequent hypocrisy allegations are.

% We randomly selected a subset of the Sustainable Initiatives data frame\footnote{\url{github.com/aapolimeno/ScrapeSustainableInitiative}} 

% for the purpose of annotating instances of accusations of hypocrisy. The Sustainable Initiatives Dataset comprises Reddit posts and comments sourced from European subreddits that contain keywords related to sustainability, such as climate change, climate crisis, and sustainability, etc. 

 % We chose to analyze the comment section instead of the original post, as we anticipate a higher likelihood of encountering allegations of hypocrisy in a deliberative space. Our unit of analysis is a comment, which can contain a single sentence or multiple sentences. 

\subsubsection{Annotation Scheme}\label{ssec:annotation}
We devised a nested annotation scheme to identify instances of hypocrisy allegations within statements (Q1), and when these are detected, the type of accusation (Q2). First, hypocrisy allegations are coded binarily: \textit{Hypocrisy Accusation/ No accusation}. A statement is considered an allegation of hypocrisy when it does at least one of the following:
\begin{itemize}
    \item Includes a direct hypocrisy accusation, such as calling someone a hypocrite or describing their actions as hypocritical (e.g., ``COP26 is the biggest hypocrisy in the world, arriving in their private jets'')
    \item Highlights a clear inconsistency or contradiction between someone's actions and their stated values, usually in a way which is negatively morally coded  (e.g., ``Leonardo DiCaprio simply doesn’t get it, protecting marine animals and flying private jets at the same time'')
    \item Employs a rhetorical device such as questioning or invoking hypothetical scenarios to indirectly accuse someone of hypocrisy (e.g., ``Shouldn’t you consider your own actions before instructing us on what needs to be done?'') 
\end{itemize}

The codebook further specifies that allegations can target individuals, institutions, or collectives. Second-hand accusations (e.g., ``Lucy Dracus said that Obama is a hypocrite'') do not constitute an accusation. Allegations can also be expressed through phrases or sayings synonymous with hypocrisy, such as ``double standard'' or ``one rule for thee, another for me''. 

In cases in which a hypocrisy accusation was detected (Q1 answered positively), the annotator proceeded to identify the type of hypocrisy (Q2). The categorization into hypocrisy types draws from Gunster’s typology \citep{Gunster2018} of climate hypocrisy discourses, which lays out a distinction between types focused on individual (individual-lifestyle outrage and personal reflective discourse) versus institutional (institutional cynicism and calls to action) behavior. Operationalizing these distinctions, we lay out the following categorization: 
\begin{enumerate}[label=\Alph*)]
    \item \textbf{Personal moral hypocrisy (PMH)}: a gap between personal behavior and professed beliefs. 
    \item[] \textit{Example:} ``You claim to care about climate change, yet you eat beef.''

    \item \textbf{Political hypocrisy (PH)}: a discrepancy between professed beliefs, values, or ideology and policy or political action.
    \item[] \textit{Example:} ``You talk about the importance of climate change but oppose nuclear power.''
    \item[] \textit{Note:} this category also includes inconsistencies between different policy positions.

    \item \textbf{Neither}: We apply this when we cannot decide between A and B, when there are reasons to choose both, or when we think that neither A or B fit.    
\end{enumerate}

When determining the type of hypocrisy, the primary consideration is the content of the targeted action, statement or position. For instance, consumer choices typically indicate personal moral hypocrisy, while explicitly political action such as voting or protesting indicates political hypocrisy. If the content is unclear, the type of actor being accused can guide the decision: accusations against nations or governments are usually political, whereas accusations against private citizens are typically personal. Accusations against specific politicians can be either, depending on the content of the allegation.  

\subsection{Climate Hypocrisy Accusations Corpus}\label{ssec:CHAC}

Our \textit{Climate Hypocrisy Accusations Corpus} (CHAC) corpus consists of 420 labeled comments. We calculate an inter-annotator agreement score of Fleiss' $\kappa$: 0.512, indicating a reasonable level of consensus among the annotators. We use majority voting to assign labels to each comment.
%and proceeded to discard the comments that did not achieve consensus. 
There are 293 comments with a majority-class assigned label (the rest did not have a majority label). However, we keep the comments without majority consensus and release it with our dataset, as recent calls for perspectivism \cite{rottger2022two,romberg2022your} have highlighted the importance of looking beyond majority consensus when it comes to complex social and argumentative concepts.
The distribution of labels is summarized in Table \ref{tab:CHAC-stats}. The source-code of the analysis\footnote{\url{https://github.com/pgarco/few-hypo}} as well as our corpus\footnote{In \url{https://huggingface.co/datasets/Myrthe/RedditEuropeanSustainabilityInitiatives}} is available online, released for non-commercial use only under CC-BY-NC licence.

\begin{table}[]
\begin{tabular}{@{}ll@{}}
\toprule
Label                    & N   \\
\midrule
Personal Moral Hypocrisy & 35  \\
Political Hypocrisy      & 35  \\
Neither                  & 2   \\
No accusation            & 221 \\
\bottomrule
\end{tabular}
\caption{Count summary label distributions of the labelled dataset}
\label{tab:CHAC-stats}
\end{table}

\section{Experimental Approach}\label{sec:experimental-setup}

Our experiments are a first attempt at using the  \textit{Climate Hypocrisy Accusations Corpus} to measure the capabilities of different currently popular and high-performing LLMs in detecting hypocrisy accusations. We use few-shot prompting, also known as in-context learning. \cite{brown2020language}. Previous research has shown that for complex social constructs, few-shot out performs zero-shot \cite{Alhindi2023}.

\subsection{Model Selection}\label{ssec:models}

Our experiments  compare two families of high-performing and currently popular LLMs that have shown promise on complex social tasks. We use two GPT series models \cite{brown2020language} and one LLama series model \cite{touvron2023llama}. 

We also purposefully chose one more closed and one more open family of models in terms of development and model access. LLama models (Meta AI) do not require payment, and its development team has openly released most of its code and training procedure, while GPT (OpenAI) does require payment and is less open in its architecture. However, almost no currently released LLM by a large technology company is fully open in its release of code, training data, and analysis (see \citet{liesenfeld2023opening} for comparing aspects of `opennness' when it comes to LLMs).\footnote{Full model description in Appendix \ref{ssec:models-app} and \ref{ssec:app-train}.}

\subsection{Prompt and Shot Selection}\label{ssec:prompt}

 We opt for a six-shot learning approach to provide the model with two robust examples per category. This provides sufficient context for each classification type without risk of overfitting. This also allows to have some control over the output format, avoiding complex parsing and streamlining the analysis of the results.

We use an iterative prompt design process using the GPT-4o  as our base model. We base our prompt on classification formats present in our literature review on fallacy detection. %and made sure to include instructions.  %and began by specifying the role the AI has to play. 
The examples we choose to include in the prompt are not in the test set and are selected to maximize model learning capabilities. This includes hypocrisy examples with complex constructions, reported speech, and rhetorical questions. We also include reasoning in our prompt, as previous literature found reasoning increases model performance, even if it is not reliably correct \cite{ye2022unreliability}.\footnote{ Full prompt is in Appendix \ref{ssec:app-prompt}} 

\section{Results}\label{sec:results}
We report our results on the dataset of 293 majority-annotated instances from our \textit{Climate Hypocrisy Accusation Corpus}. To ensure consistency and accuracy in our analysis, we use a systematic parsing process for the strings generated by each model. We standardize the formatting of the outputs to eliminate any discrepancies in punctuation, capitalization, and spacing, using regex patterns to detect and correct common formatting issues, as well as custom scripts designed to handle unique idiosyncrasies of each model's outputs.

\subsection{Overall Results}

\begin{figure}
    \centering
    \includegraphics[width=0.9\columnwidth]{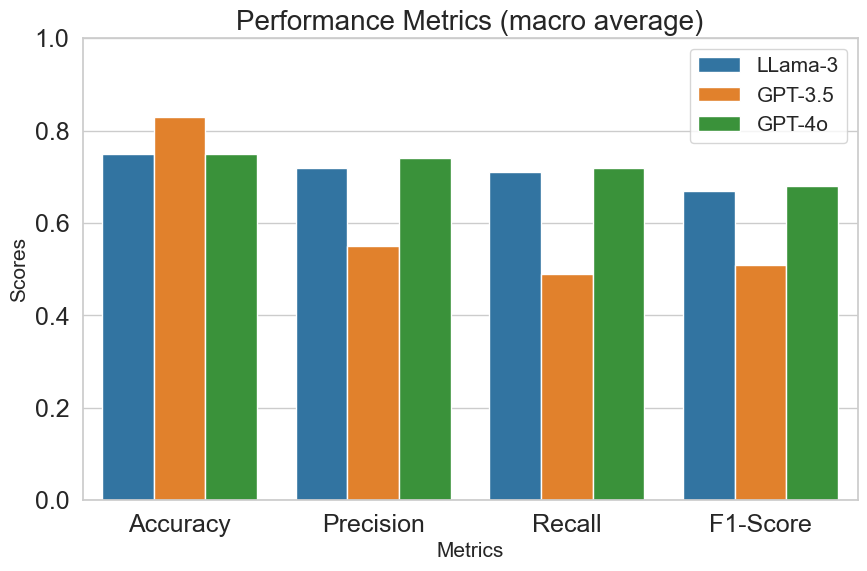}
    \caption{Bar graph comparing result metrics of LLM performance, from left to right we see LLama-3 (blue), GPT-3.5 (orange), and GPT-4o  (green), grouped by accuracy (first group), precision (second group), recall (third group), and F1-score (last group).  }
    \label{fig:bar-metrics}
\end{figure}

\begin{figure}
    \centering
    \includegraphics[width=0.9\columnwidth]{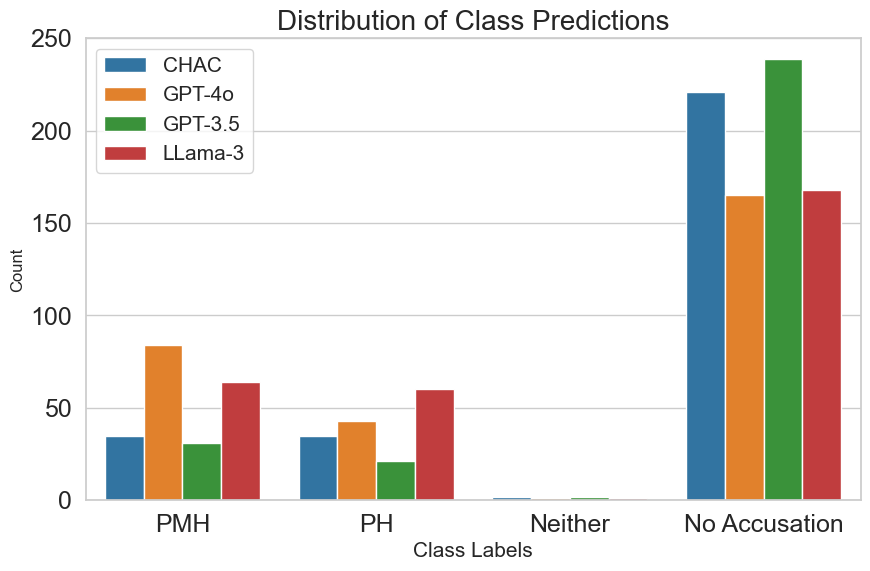}
    \caption{Bar graph comparing prediction and real labels distribution: from left to right we see CHAC dataset (blue), GPT-4o  (orange), GPT-3.5 (green) and Llama-3 (red) grouped by class label: PMH (first group), PH (second group), Neither (third group), and No accusation (last group).}
    \label{fig:bar-labels}
\end{figure}

The classification results can be seen in 
Table \ref{tab:clf-results} and Figure \ref{fig:bar-metrics}. LLama-3 and GPT-4o  both perform relatively well, significantly outperforming GPT-3.5 overall. In terms of accuracy, all models make predictions that are correct at least 75\% of the time, with GPT-3.5 actually leading the way (83\%). However, this high accuracy in GPT-3.5 appears to be an artefact of the imbalance between categories. As Table \ref{tab:dist-results} and Figure \ref{fig:bar-labels} show, the ``No accusation'' label is far more prevalent than the other categories, and GPT-3.5 does a better job at predicting it, while GPT-4o  and LLama-3 under-predict this class but are better in detecting the two hypocrisy classes.

\begin{table}[]
\begin{tabular}{@{}lllll@{}}
\toprule
        & Acc  & Prec & Recall & F1 \\
        \midrule
LLama-3 & 0.75 & 0.72 & 0.71 & 0.67 \\ 
GPT-3.5 & 0.83 & 0.55 & 0.49 & 0.51  \\
GPT-4o   & 0.75 & 0.74 & 0.72 & 0.68 \\
\bottomrule
\end{tabular}
\caption{Classification results on the 293 examples labelled for hypocrisy. Results of LLama-3, GPT-3.5, and GPT-4o  in accuracy, precision, and recall. }
\label{tab:clf-results}
\end{table}

\subsection{Sub-class Prediction}
The important difference between the models lies in the prediction of the two hypocrisy classes, reflected in both precision and recall. While GPT-3.5 managed both tasks roughly half of the time, LLama-3 and GPT-4o  both succeeded at both tasks above 70\% of the time. 

Results of the sub-class predictions are visible in Table \ref{Classification results for GPT-4o } for GPT-4o , Table \ref{Classification results for GPT-35-turbo-1025} for GPT-3.5, and Table \ref{Classification results for Llama-3-70b-chat-hf} for Llama-3. All models perform worse when identifying both Personal Moral Hypocrisy (PMH) (F1 scores between 0.67 and 0.63) and Political Hypocrisy (PH) (F1 scores between 0.46 and 0.54) than in identifying ``no accusation'' (0.91). All over-predict the different subtype labels, and are worst at identifying accusations not falling under either subtype, though these accusations are very rare. Overall,  LLama-3 and GPT-4o  respectively have nearly-identical macro-averaged F1 scores of 0.67 and 0.68 over all classes. 

We view these scores as a good benchmark for complex hypocrisy accusation detection. The LLama-3 and GPT-4o  models show potential for identification of hypocrisy accusations and classification of specific hypocrisy types. 

\begin{table}[h]

\resizebox{\columnwidth}{!}{%
\begin{tabular}{@{}lllll@{}}
\toprule
              & precision & recall & f1-score & support \\ \midrule
No accusation & 0.98      & 0.73   & 0.84     & 221     \\
Personal Moral Hypocrisy           & 0.60      & 0.74   & 0.67     & 35      \\
Political Hypocrisy            & 0.38      & 0.91   & 0.54     & 35      \\
Neither       & 1.00      & 0.50   & 0.67     & 2       \\ \midrule
accuracy      &           &        & 0.75     & 293     \\
macro avg     & 0.74      & 0.72   & 0.68     & 293     \\
weighted avg  & 0.87      & 0.75   & 0.78     & 293     \\ \bottomrule
\end{tabular}%
}
\caption{Multiclass classification results for GPT-4o  on the 293 examples with a majority label for hypocrisy acccusations}
\label{Classification results for GPT-4o }
\end{table}

\begin{table}[h]
\resizebox{\columnwidth}{!}{%
\begin{tabular}{@{}lllll@{}}
\toprule
              & precision & recall & f1-score & support \\ \midrule
No accusation & 0.87      & 0.95   & 0.91     & 221     \\
Personal Moral Hypocrisy  & 0.71      & 0.63   & 0.67     & 35      \\
Political Hypocrisy & 0.62      & 0.37   & 0.46     & 35      \\
Neither       & 0.00      & 0.00   & 0.00     & 2       \\ \midrule
accuracy      &           &        & 0.83     & 293     \\
macro avg     & 0.55      & 0.49   & 0.51     & 293     \\
weighted avg  & 0.82      & 0.83   & 0.82     & 293     \\ \bottomrule
\end{tabular}%
}
\caption{Multiclass classification results for GPT-3.5-turbo-1025 on the 293 examples with a majority label for hypocrisy acccusations.}
\label{Classification results for GPT-35-turbo-1025}
\end{table}

\begin{table}[h]
\resizebox{\columnwidth}{!}{%
\begin{tabular}{@{}lllll@{}}
\toprule
              & precision & recall & f1-score & support \\ \midrule
No accusation & 0.97      & 0.74   & 0.84     & 221     \\
Personal Moral Hypocrisy          & 0.50      & 0.86   & 0.63     & 35      \\
Political Hypocrisy & 0.41      & 0.74   & 0.53     & 35      \\
Neither       & 1.00      & 0.50   & 0.67     & 2       \\ \midrule
accuracy      &           &        & 0.75     & 293     \\
macro avg     & 0.72      & 0.71   & 0.67     & 293     \\
weighted avg  & 0.85      & 0.75   & 0.77     & 293     \\ \bottomrule
\end{tabular}%
}
\caption{Multiclass classification results for Llama-3-70b-chat-hf on the 293 examples with a majority label for hypocrisy acccusations.}
\label{Classification results for Llama-3-70b-chat-hf}
\end{table}

\begin{table}[]
\resizebox{\columnwidth}{!}{%
\begin{tabular}{@{}lcccc@{}}
\toprule
Label         & GPT-3.5 & GPT-4o  & LLama-3 & CHAC \\
\midrule
Personal Moral Hypocrisy          & 31    & 43  &  60  &  35  \\
Political Hypocrisy & 21    & 84  &  64  &  35  \\
Neither       & 2     & 1   &  1   &  2   \\
No accusation & 239   & 165 &  168 &  221 \\
\bottomrule
\end{tabular}%
}
\caption{Distribution of Class Prediction on the 293 examples with a majority label for hypocrisy acccusations.}
\label{tab:dist-results}
\end{table}

\subsection{Error Analysis}\label{ssec:error-anal}
We perform an error analysis to see whether there are error patterns relating to model type in hypocrisy accusation detection. Overall, we do not find a connection between error type and model family. Each model makes distinct errors.\footnote{Confusion Matrices for all models are in Appendix \ref{ssec:app-error}.} 

%We see a persistence in misclassification even when there is a reasoning component asked in the prompt, which we expected based on literature would lead to a reduction of errors. 

Broadly, we see three types of error in our results. Firstly, the LLMs predict False Positives, which we refer to as \textbf{hypocrisy accusation hallucinations}, often accompanied by a false reasoning in the explanation of the labelling decision generated by the model. A second common error is misclassification of subtype: a correct identification of an accusation, but an incorrect classification of the accusation type. Less common are False Negatives, where a hypocrisy accusation is not found where there is one. Below we discuss each type of error in the different models.

\paragraph{False Positives: Accusation Hallucination}
While all models sometimes identify hypocrisy accusations where there are none, we find different patterns between the models. GPT-4o  and LLama-3 overpredict accusations where there are none, predicting 59 and 58 false positives, respectively. GPT-3.5, meanwhile, hallucinated accusations far less, with only 12 cases of false positives. 

Further investigating false positives, we find the cases of hypocrisy accusation hallucination that include the regex pattern \verb|hypocr*| to determine if the presence of the \textit{mention} of hypocrisy could explain the models' over-prediction. Not every mention of the word hypocrisy actually contains a hypocrisy accusation, but it may be a common confusion for models. We find that 37 out of 59 (62\%) of GPT-4o's false positives contain the pattern \verb|hypocr*|. For the Llama-3 model, 31 cases (52\%), and for the GPT-3.5 errors only 9 cases out of 28 (32\%) match the regex pattern. Thus, we observe that GPT-3.5 is better at distinguishing a hypocrisy \textit{accusation} from the \textit{mention} of hypocrisy than other models. See the example below.\footnote{Appendix \ref{App:error_analysis} has examples of the different types of errors.}

\begin{itemize}
  \item[] \textit{“What will happen is when it is declared a national emergency, the right will call the left hypocrites for only caring because it was Trump. When in reality, the right’s emergency is a baseless claim and climate change is \emph{fucking} real.”}

  Human label: \textbf{Not an accusation}

  Llama-3 predicted label: \textbf{Political Hypocrisy}
  
  Error: This is a case of the model classifying a second-hand accusation (and a hypothetical one at that) as a hypocrisy allegation. We would not wish to classify this as an allegation: the comment is reporting on this purported accusation in order to refute it. 

    \item[] \textit{“By that logic basically everyone who wants to stop climate change is a hypocrite.”}

  Human label: \textbf{Not an accusation}

  GPT-4o  predicted label: \textbf{Personal Moral Hypocrisy}
  
 Error: We understand this comment to mention a hypocrisy accusation critically, presumably responding to a hypocrisy allegation by pointing out how its reasoning leads to a conclusion that is \textit{prima facie} absurd. The GPT-4o model seemed to simply take the comment at face value, reasoning that ``The commenter is suggesting that anyone who advocates for stopping climate change is a hypocrite''. This points to the complexity of understanding nuanced concepts such as irony or sarcasm. 
  \end{itemize}

\paragraph{False Negatives}
GPT-3.5 misclassifies 28 comments as false negatives – not correctly identifying an accusation. False negatives, then, are the main source of errors for GPT-3.5.

We again look into whether there are any confusions between the \textit{mention} of the word 'hypocrisy' and an \textit{accusation} of hypocrisy. We find that GPT-4o  did not label any comment that contains the regex pattern \verb|hypocr*| as \textit{not} being accusations when these were positive cases, while for Llama-3 we find 3 cases of such false negatives. 

GPT-3.5 labeled 16 false negative comments containing the pattern. This indicates that GPT-3.5 is not relying on the presence of \verb|hypocr*| related words to label comments, compared to the two other larger models, which is a surprising finding. However, as we see, this did not lead to a better overall performance for GPT-3.5 in detection hypocrisy accusations.

% \noindent \textbf{Examples of False Positives}

% \begin{enumerate}
%   \item \textit{“What will happen is when it is declared a national emergency, the right will call the left hypocrites for only caring because it was Trump. When in reality, the right’s emergency is a baseless claim and climate change is \emph{fucking} real.”}

%   Human label: \textbf{Not an accusation}

%   Llama-3 predicted label: \textbf{Political Hypocrisy}
  
%   This is a case of the model classifying a second-hand accusation (and a hypothetical one at that) as a hypocrisy allegation. We would not wish to classify this as an allegation, especially as the comment is reporting on this purported accusation in order to refute it. 
  
%   \item \textit{“By that logic basically everyone who wants to stop climate change is a hypocrite.”}

%   Human label: \textbf{Not an accusation}

%   GPT-4o  predicted label: \textbf{Personal Moral Hypocrisy}
  
%   While this comment makes an explicit suggestion of hypocrisy, we understand it to be doing so critically, presumably responding to a hypocrisy allegation by pointing out how its reasoning leads to a conclusion that is prima facie absurd. Two of the models concurred, but the GPT-4o model seemed to simply take the comment at face value, reasoning that "The commenter is suggesting that anyone who advocates for stopping climate change is a hypocrite". This points to the complexity of understanding nuanced concepts such as irony or sarcasm. 
% \end{enumerate}

\paragraph{Subtype Misclassification} Aside from confusing non-accusations with accusations, the models also show confusion in the subtypes of hypocrisy. 
We find that these errors are not consistent across models. In the case of \textit{Personal} hypocrisy being labeled as \textit{Political}, the biggest confusion comes from GPT-4o  (22\%). While Llama-3 has the opposite error: \textit{Political} is labeled as \textit{Personal} (17\%). Again, our findings indicate that the smaller model, GPT-3.5, does not confuse types of hypocrisy as much, with only 4 instances of \textit{Political} predicted as \textit{Personal}, and one single case of the opposite. While GPT-3.5 under-predicted the hypocrisy accusations (having more false negatives), it was better at distinguishing between the hypocrisy classes once they were labeled as accusations. 

\begin{itemize}
  \item[] \textit{“It's James Shaw, the biggest hypocrite out there when it comes to travel. Don't expect a realistic response.”}

  Human label: \textbf{Personal Moral Hypocrisy}

  GPT-4o  predicted label: \textbf{Political Hypocrisy}

  Error: The model appears to conclude that James Shaw, the former New Zealand Minister for Climate Change, is a political figure. However, we have noted that accusations against specific politicians can be personal or political depending on their content, and would classify personal travel as a consumer choice and thus an example of personal hypocrisy. However, we recognize this is something of a gray area, and that the model and human coders disagree on it.  
\end{itemize}

% \vspace{0.2cm}

% \noindent \textbf{Examples of Misclassifications}

% \begin{enumerate}

%   \item \textit{“It's James Shaw, the biggest hypocrite out there when it comes to travel. Don't expect a realistic response.”}

%   Human label: \textbf{Personal Moral Hypocrisy}

%   GPT-4o  predicted label: \textbf{Political Hypocrisy}

%   The model appears to based its reasoning on the fact that James Shaw, the former New Zealand Minister for Climate Change, is a political figure. However, we have noted that accusations against specific politicians can be personal or political depending on their content, and would classify personal travel as a consumer choice and thus an example of personal hypocrisy. However, we recognize this is something of a gray area and it perhaps unsurprising that the model and human coders disagree on it.  

%   \item \textit{“Conservatives are bollocks but let me tell you how good their policies actually are! Hypocrite much?”}

%   Human label: \textbf{Political Hypocrisy}

%   Llama-3  predicted label: \textbf{Personal Moral Hypocrisy}

%   The comment points to a clear inconsistency between one's political belief (conservatives are bad) and action (praising their policies). However, the GPT-4o  model appears to interpret on the fact that the accusation appears directly at another speaker as evidence it is a case personal moral hypocrisy. 
% \end{enumerate}

We want to find the source of confusion of classes, and investigate if the mention of political figures or events leads to mislabeling. A qualitative analysis reveals that all subtype misclassification cases for GPT-4o  include references to political figures or events. In this case, correct annotation of subtype (political hypocrisy or not) requires the model to establish whether the comment has a political context. The errors in LLama-3 and GPT-3.5 of identifying \textit{Political Hypocrisy} as \textit{Personal Moral Hypocrisy} all have references to political figures or events. All four subtype errors by GPT-3.5 are a subset of the error cases made by LLama-3. However, we cannot conclude that references to political figures or events are the source of subclass error, but it is possible that the model is unable to correctly identify mentions as political.

In summary, our error analysis shows some interesting patterns. GPT-3.5, our least successful model overall, is conservative in labelling comments as hypocrisy accusations, but therefore also has less false positives than the other models. We furthermore find this model is less confused by the mention of the word `hypocrisy'. The other two models, LLama-3 and GPT4o, are better at labelling comments, but also confuse the two subtypes of hypocrisy more. Overall, political hypocrisy seems hard to identify for all models, and the models struggle with identifying a political context from the mention of political actors.

\section{Discussion}\label{sec:discussion}
Our results provide more insight into the complexities of the hypocrisy accusation as a construct. Specifically, we find different model strengths (see Section \ref{sec:modeldif}) and specifics of different types of hypocrisy accusations (Section \ref{sec:complexconstr}).

\subsection{Model Difference}\label{sec:modeldif}
Our results indicate that the most recent models, including GPT-4o, are considerably better than the earlier GPT3.5. This indicates that the current development of instruction-tuned models is one where improvement also means being better at detecting and annotating a complex construct like hypocrisy accusations. Newer models seem to be better at identifying nuanced context. One possible reason for this may be  more recently updated training data e.g. new politicians or political events, required to fully understand hypocrisy accusations. %the relevance of new data on new topics in the training data, given that context matters in hypocrisy detection. 
However, due to lack of model training transparency, this cannot be verified.

Additionally, the acceptable results of both LLama-3 and GPT-4o  are worth considering in light of the former being more open source and freely available, and the latter a more closed model requiring payment. Given their similar performance, the more open model offers clear advantages to researchers in social science. More open models are better for science: these are more reproducible, understandable and allow researchers to not be dependent on paying a third party \cite{liesenfeld2023opening}. While none of our tested models are completely open in all aspects (training data, code, and openly accessible), it is useful for social scientists  to know that Llama is a LLM that is easily accessible, able to analyze a complex construct such as a climate hypocrisy accusation, and not requiring third-party payment.

\subsection{Results in the Context of Other Research}\label{sec:complexconstr}
The results also suggest that hypocrisy detection, as most logical fallacies, is a complex task. A pattern-matching approach would consider mentions of `hypocrite' as positive cases, but – as our experiments illustrate – such accusations also occur indirectly, and their detection is complicated by reported speech, sarcasm, rhetorical questions, and other devices in online debates. %In line with our hypothesis, 
We observe that using LLM in-context prompting, detection of hypocrisy accusations achieves decent results. However, compared to other classification tasks for social science constructs \cite{lim2024evaluation}, using LLMs to classify accusations of hypocrisy is below expected model capabilities. 

Our results (Macro F1 = 0.68 for GPT-4o ) see a performance gap with the fallacy detection on other fallacies reported in earlier work, e.g. F1 = 0.76 for fallacy accusation detection in \citet{sahai2021breaking}, and 0.81 in \citet{Habernal2018} for detecting \textit{ad hominem} attacks. However, these earlier papers did not distinguish hypocrisy accusations from other fallacies, and this narrowing of the concept could lead to more difficulty for our models. 

Other literature also reports a more mixed performance of LLMs (especially when compared to fine-tuned Transformer models) for fallacy detection. \citet{ruiz2023detecting} report a F1 score of 0.79 by a fine-tuned Roberta model for two fallacy argument datasets with classes such as \textit{ad hominem attacks} and \textit{appeal to majority}, and in contrast a 0.56 F1 score for GPT-4o on these datasets. This paper also reports a lower performance of GPT-4o on the \textit{ad hominem} fallacy class. Potentially, this could be because of a connection to hypocrisy accusations, which (as we have established) are difficult, and are often forms of \textit{ad hominem} arguments, criticizing the rival personally instead of their positions.

Our results indicate that hypocrisy accusation is an interesting concept that deserves its own task, as well as benchmark datasets outside of more common fallacy datasets. %- which often do not include hypocrisy accusation as a separate class. 
Moreover, these results prove the usefulness of breaking down complex constructs into sub-categories: we found that detecting the hypocrisy/no hypocrisy distinction is relatively easy (e.g., reaching F1 > 0.80 for all models), while the subclass Political Hypocrisy is much harder to detect, showing F1s <= 0.50s for all models. The subclass that may be especially relevant for political analysis, attacks on political actions or views, is not well-detected by LLMs.

% While previous research has focused on fallacy detection as a whole, these results demonstrate that certain types of fallacies are especially hard to detect, and need further research. These results should be taken into account when building models for policy or political discourse, as they demonstrate the importance of specific constructs, which can greatly improve results if they are further broken into categories.

\section{Conclusions}\label{sec:conclusions}
Hypocrisy accusations are central to increasingly polarized, cross-ideological online interactions. Despite recent research on detecting argument fallacies, hypocrisy accusations remain under-researched and are often a small sub-class in argument fallacy datasets. We define hypocrisy accusation detection as an individual NLP task and create an annotation scheme where we identify subclasses of hypocrisy accusation: personal moral hypocrisy versus political hypocrisy. We present a dataset, the \textit{Climate Hypocrisy Accusation Corpus} (CHAC), consisting of 420 reddit comments, annotated by six experts. %To our knowledge, CHAC is the first corpus that classifies different types of hypocrisy accusations for computational analysis. 
Using our dataset, we compare three different instruction-tuned models (GPT-4o , GPT-3.5, and Llama-3) in a six-shot setting for detecting hypocrisy accusations. The different models have different strengths, but overall perform with a macro F1 class of around 0.80, and show that Llama, as a more open model than the GPT-family and one not requiring payment, can perform on par for hypocrisy accusation detection with the less open GPT model that requires payment. LLMs are capable of detecting accusations with a binary distinction, but we identify room for improvement when it comes to the different accusation types. Models are somewhat worse at detecting political hypocrisy than personal moral hypocrisy, which could have implications for social science research.
% Our results show that LLMs are capable of detecting accusations with a binary distinction, but there's room for improvement when it comes to accusation types. All in all, our work shows that, within the limitations discussed, large language models can be used to detect complex concepts such as accusations of hypocrisy in online climate change debates. 
% While there may not be perfect accuracy between the models and human coding, we have shown that LLMs have made significant progress and offer great potential for analyzing large datasets to uncover complex semantic concepts that would otherwise be missed. 
%We hope to continue our work in the future by increasing our data sample and extending our case selection to other policy areas that are arenas of contestation.

\newpage

\section*{Limitations}\label{sec:limitations}
As with all research, this paper has some limitations. We identify four sources of limitations.

\paragraph{Data Annotation}
First, the data annotation process could have led to higher inter-annotator agreement score, which, research has shown, is detrimental to achieving high results in computational modelling. However, considering this is a complex theoretical construct, we are satisfied with this first limited result. 

\paragraph{Debate Context}
Second, the validity of the results should be understood as pertaining especially to climate change discourse; As we have described in the paper, hypocrisy relies on the understanding of a contrast between two events, usually a professed belief and an action. These actions and beliefs often need to be understood in context in order to be understood as (allegedly) inconsistent. Hence, we expect identification of hypocrisy accusations to be somewhat dependent on an understanding of both the factual reality of a topic as well as the social context in which it is discussed. This paper analyzes the climate change debate, and the tool's relevance to other fields requires further study and, potentially, training. 

\paragraph{Geographical, Linguistic, and Cultural Context}
Additionally, we acknowledge that our paper is focused on European debates around climate change, in a single high-resource language, English \cite{bender2019benderrule}. The results we find depend on the data the models were trained on, and as such we expect that non-European debates and debates in low-resource languages will probably produce results that are not as high.

\paragraph{Political Context}
Lastly, LLMs are not without its issues for social science analysis: these models display political worldviews \cite{ceron2024beyond}. When analyzing different political contexts (e.g. one more conservative than European climate debates), the results could therefore differ. While adding these results is beyond the scope of the current paper, it is important to keep this in mind for future work. 

\section*{Ethics Statement}\label{sec:ethics}
The data used in this project was scraped from Reddit in December 2022 with the PushShift API, before Reddit's PushShift API restrictions were enforced in April 2023, ensuring compliance with the platform's terms of service at the time. We remove any personal identifying information such as usernames from the data. We also ensure the data is released for non-commercial use only. This is also in-line with Reddit users' concern of their data being used for training commercial LLMs or other technology. 

Furthermore, some comments reflect personal opinions that are not in-line with the established scientific consensus on climate change. While these opinions are valuable for understanding public sentiment and discourse, we do not endorse any misinformation or scientifically inaccurate statements present in the dataset. Our goal is to analyze these discussions to better understand the dynamics of public discourse on climate change, promoting more effective strategies to engage with the public, address misconceptions, and promote scientifically accurate information.

\section*{Acknowledgements}\label{sec:ac}
The dataset used in this research was funded by a research voucher grant for the project \textit{Reasons for online (dis)trust in sustainable initiatives} awarded to Myrthe Reuver and Ana Isabel Lopes by the Network Institute at the Vrije Universiteit Amsterdam. Myrthe Reuver was also funded by the Netherlands Organisation for Scientific Research (NWO) through the \textit{Rethinking News Algorithms} project (grant nr: 406.D1.19.073). Paulina Garcia Corral was funded by the DFG (EXC number 2055 – Project number 390715649, SCRIPTS).

A preliminary version of this paper was presented at the hackathon at the International Communication Association (ICA) 2023 conference in Toronto, where this project was initiatied and where all authors participated, and another preliminary version was presented at CompText 2024 in Amsterdam. We would like to thank the ICA hackathon organizers and participants as well as all reviewers, whose comments improved this paper.

We used Github Copilot and ChatGPT to assist in writing our experimental code, and editing code snippets for data processing and data analysis. We used ChatGPT to assist in writing sections of this manuscript. This assistance, in accordance with the ACL Ethics Policy, was solely in dealing with the language of the paper rather than in producing new content or new ideas. All final writing is ultimately done by the authors, who are responsible for it.

\newpage
\bibliography{custom}
\bibliographystyle{acl_natbib}

\newpage\phantom{blabla}

\appendix

\section*{Appendix}\label{sec:appendix}

\section{Implementation Details}\label{ssec:models-app}
\subsection{Model Details}
\begin{itemize}
    \item GPT-3.5-turbo-0125: This is an OpenAI GPT series \cite{brown2020language} model, optimized for efficient and cost-effective performance in conversational AI tasks, providing advanced language understanding and generation capabilities. This is a closed-source model.
    \item GPT-4-turbo: An enhanced version of OpenAI's GPT-4, this model offers improved speed and performance for complex language processing tasks, making it ideal for both conversational agents and other sophisticated AI applications. This is a closed-source model.
    \item Llama-3-70b-instruct\footnote{\url{https://huggingface.co/meta-llama/Meta-Llama-3-70B-Instruct}}: Developed by Meta AI \cite{llama3}, this is a language model with 70 billion parameters, designed for high-quality conversational AI, capable of understanding and generating human-like text in diverse contexts. This is a non-proprietary model, i.e., it does not require payment. 
\end{itemize}

\subsection{Prompt}\label{ssec:app-prompt}
\textbf{System:} \\
You are an advanced classification AI. Your task is to labels Reddit comments following the instructions below: \\
**Instructions* \\
Accusations of hypocrisy are when someone accuses someone or something of being a hypocrite because of a moral contradiction between what they profess and their actions or words. Accusations of hypocrisy are sometimes direct such as ``You are a hypocrite'' and sometimes are understood via context, by highlighting a connoted inconsistency between what people profess or claim their believe and what people actually do. An example would be criticizing a celebrity for using a private jet while publicly claiming the need for climate action.\\
There are different types of hypocrisy accusations, such as accusing someone of personal moral hypocrisy or accusing someone of political hypocrisy. The primary consideration to decide the accusation type is the content of the targeted action, statement or position. For instance, consumer choices typically indicate personal moral hypocrisy, while explicit political action such as voting or protesting indicates political hypocrisy. If the content is unclear, the type of actor being accused can guide the decision: accusations against nations or governments are usually political, whereas accusations against private citizens are typically personal. \\
**Classification steps* \\
1. You will analyze a Reddit Comment and evaluate if there is an accusation of hypocrisy. \\
2. You will make sure to consider contextual factors, implications, and rhetorical devices to make sure things such as irony or double meaning are taken into account. \\
3. You will assign a label to the comment: ``Not an accusation'' or ``Hypocrisy accusation''. \\
4. If the label is ``Hypocrisy accusation'', then you will analyze what type of accusation it is: \\
1) ``Personal moral hypocrisy'': a gap between personal behavior and professed beliefs \\
2) ``Political hypocrisy'': a gap between professed beliefs, values, or ideology and policy or political action. \\
3) ``Neither'' \\
5. Then you will assign a second label to the hypocrisy accusations: ``Personal moral hypocrisy'', ``Political hypocrisy'' or ``Neither''. \\
Choose the most appropriate label, and remember to consider all the rhetorical devices. \\
\textbf{User:} \\
Label the next comment: \\
Not being vegan seems kinda hypocritical to me, if you're serious about climate change and have looked into it then you'd know that it's honestly the easiest way we can make a big impact as consumers. Vegan activists aren’t hurting anyone or being violent. Meanwhile, farming is directly causing pain, suffering, and death to animals for profit. That’s the real difference. \\
\textbf{Assistant:} \\
Label: Hypocrisy accusation - Personal moral hypocrisy \\
Reasoning: The commenter is directly stating that it is hypocritical to care about climate change and not be vegan. This identifies a perceived gap between the professed belief in fighting climate change and the personal behavior of not adopting a vegan lifestyle. The hypocrisy accusation here is based on an individual's personal moral stance (caring about climate change) not aligning with their personal behavior (not being vegan). \\
\textbf{User:} \\
Label the next comment: \\
For real?? Are you saying their hypocrites because they protest carbon mining? I've seen the mines firsthand they're wrecking nature, displacing people, and demolishing villages just for that dirty, low-energy rock. Get it right, it's the most disgusting and destructive form of electricity generation ever. \\
\textbf{Assistant:} \\
Label: Not an accusation \\
Reasoning: The invocation of hypocrisy in the second sentence (“Hypocrites” because…?”) is not itself an accusation, but a rhetorical questioning of such an allegation, one which the commenter subsequently answers negatively (“Smh”, meaning “shaking my head”, i.e. expressing incredulity at the accusation). The commenter goes on to express frustration and criticism towards the use of lignite for electricity generation, highlighting its environmental impact and the availability of other energy sources. Bottom line: the commenter is defending someone from a hypocrisy accusation. \\
\textbf{User:} \\
Label the next comment: \\
Climate activists think they’re on a moral high horse, but they’re just making life tougher for everyone else. We’ve got jobs to get to, and they claim they’re saving the future but end up screwing up our daily commutes. No one feels sorry for them—that's not how you protest effectively. \\
\textbf{Assistant:} \\
Label: Hypocrisy accusation - Political hypocrisy \\
Reasoning: The commenter is accusing climate activists of claiming to care about the future of humanity while making daily life harder for others, specifically through disrupting commutes. This highlights a perceived inconsistency between their professed goal (caring for humanity's future) and their actions (disrupting daily life). The hypocrisy accusation is focused on the contradiction between the activists' political goal (caring about the future of humanity) not aligning with the direct results results their behavior (disrupting commutes and making life harder for others). \\
\textbf{User:} \\
Label the next comment: \\
Ugh, this ``Green'' parties are pathetic. They choose Brown coal over nuclear energy, but won't shut up about air pollution. Get a grip. \\
\textbf{Assistant:} \\
Label: Hypocrisy accusation - Political hypocrisy \\
Reasoning: The commenter is highlighting a (purported) inconsistency between what the Greens profess to care about (\"crying around about air pollution\") and what they actually have done (\"preferred Brown coal over nuclear energy\"). In addition, this alleged contrast carries a clear negative connotation (\"crying around\"). The inconsistency being alleged is between professed values (caring about air pollution) and political action (supporting coal over nuclear energy). \\
\textbf{User:} \\
Label the next comment: \\
I read only that he was accused of being not only a war criminal, but also a hypocrite. \\
\textbf{Assistant:} \\
Label: Not an accusation \\
Reasoning: While the comment does report on an explicit hypocrisy accusation, this is an instance of reported speech. We count only hypocrisy allegations made or endorsed by the commenter, not descriptions of accusations made by someone else. \\
\textbf{User:} \\
Label the next comment: \\
I agree that, if it's there, people need to use trains instead of flying.\\
However, what does it matter if it's hypocritical? Just because *they* are hypocrites and chose to fly, doesn't mean what they say is untrue. The world is complicated, grow up. \\
\textbf{Assistant:} \\
Label: Hypocrisy accusation - Personal moral hypocrisy \\
This comment it likely responding to a hypocrisy accusation. Even though the commenter seeks to defend the people in question from the criticism inherent in such an accusation, they do so by conceding that the hypocrisy allegation is likely true (\"just because they are hypocrites\") - even if they contend this is not in fact so bad. While hypocrisy allegations in general carry a negative connotation, this is an unusual case in which such an allegation is explicitly made (or conceded) while attempting to divorce it from the (usual expected) negative charge. The inconsistency being alleged in this allegation stems from personal lifestyle choices - choice of commute, i.e. not taking a train. \\
\textbf{User:} \\
Label the next comment: \verb|REDDIT COMMENT|\\

\pagebreak

\subsection{Training strategy, model parameters, and other resources}\label{ssec:app-train}

For data processing we used Pandas. To test our prompt and run inference we used the OpenAI API for the GPT models, OpenAI provides their own processing equipment and is not specified. For the Meta AI model, we used TogetherAI API for inference. We used 1 Nvidia A100 GPU for our experiment. We used all default model parameters and set the generation parameters to restrict creativity (temperature = 0, max\_tokens=200, top\_p=1, frequency\_penalty=0, presence\_penalty=0). To ensure reproducibility, each model's output was subjected to identical preprocessing and post-processing steps. To analyze classification results, we used the ``classification\_report'' function from Scikit-learn.

\section{Error Analysis Figures}\label{ssec:app-error}

\begin{figure}[h!]
    \centering
    \includegraphics[width=0.9\columnwidth]{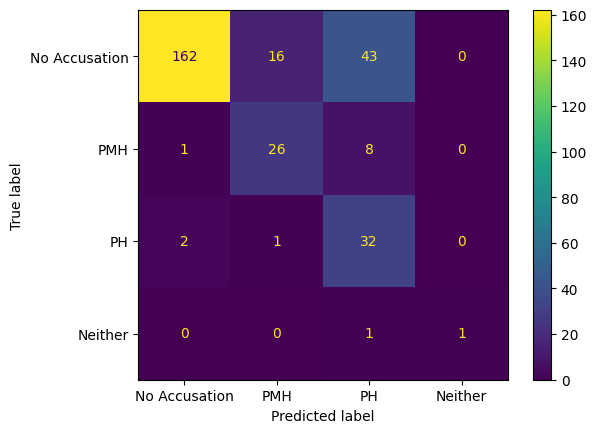}
    \caption{Confusion Matrix for predictions of GPT-4o .}
    \label{fig:gpt4-cm}
\end{figure}

\begin{figure}[h!]
    \centering
    \includegraphics[width=0.9\columnwidth]{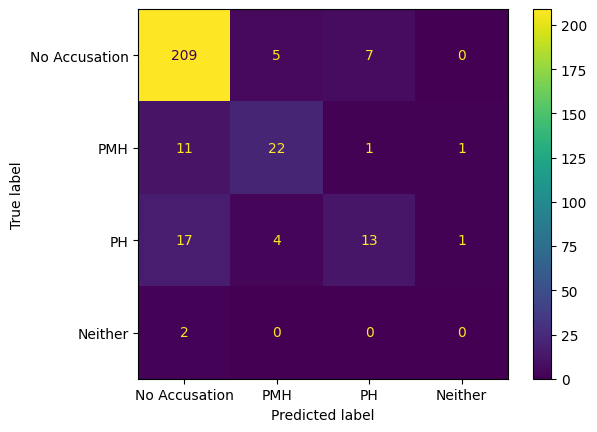}
    \caption{Confusion Matrix for predictions of GPT-3.5.}
    \label{fig:gpt35-cm}
\end{figure}

\begin{figure}[h!]
    \centering
    \includegraphics[width=0.9\columnwidth]{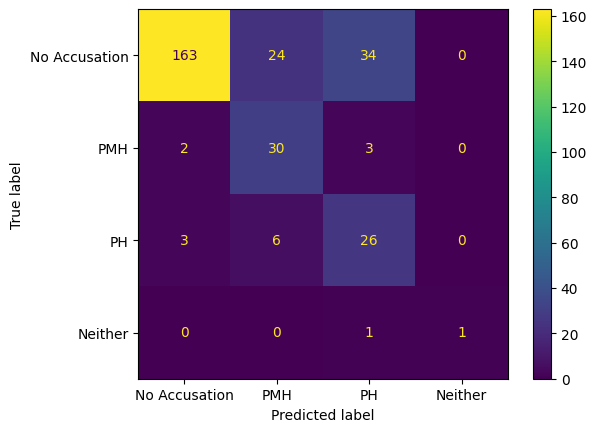}
    \caption{Confusion Matrix for predictions of Llama-3 70B.}
    \label{fig:llama-cm}
\end{figure}

\pagebreak
\section{Examples Error Analysis}\label{App:error_analysis}

\noindent \textbf{Examples of Misclassifications}

\begin{enumerate}

  \item \textit{“It's James Shaw, the biggest hypocrite out there when it comes to travel. Don't expect a realistic response.”}

  Human label: \textbf{Personal Moral Hypocrisy}

  GPT-4o  predicted label: \textbf{Political Hypocrisy}

  The model appears to based its reasoning on the fact that James Shaw, the former New Zealand Minister for Climate Change, is a political figure. However, we have noted that accusations against specific politicians can be personal or political depending on their content, and would classify personal travel as a consumer choice and thus an example of personal hypocrisy. However, we recognize this is something of a gray area and the model and human coders disagree on it.  

  \item \textit{“Conservatives are bollocks but let me tell you how good their policies actually are! Hypocrite much?”}

  Human label: \textbf{Political Hypocrisy}

  Llama-3  predicted label: \textbf{Personal Moral Hypocrisy}

  The comment points to a clear inconsistency between one's political belief (conservatives are bad) and action (praising their policies). However, the GPT-4o  model appears to interpret on the fact that the accusation appears directly at another speaker as evidence it is a case personal moral hypocrisy. 
\end{enumerate}

\noindent \textbf{Examples of False Positives}

\begin{enumerate}
  \item \textit{“What will happen is when it is declared a national emergency, the right will call the left hypocrites for only caring because it was Trump. When in reality, the right’s emergency is a baseless claim and climate change is \emph{fucking} real.”}

  Human label: \textbf{Not an accusation}

  Llama-3 predicted label: \textbf{Political Hypocrisy}
  
  This is a case of the model classifying a second-hand accusation (and a hypothetical one at that) as a hypocrisy allegation. We would not wish to classify this as an allegation, especially as the comment is reporting on this purported accusation in order to refute it. 
  
  \item \textit{“By that logic basically everyone who wants to stop climate change is a hypocrite.”}

  Human label: \textbf{Not an accusation}

  GPT-4o  predicted label: \textbf{Personal Moral Hypocrisy}
  
  While this comment makes an explicit suggestion of hypocrisy, we understand it to be doing so critically, presumably responding to a hypocrisy allegation by pointing out how its reasoning leads to a conclusion that is prima facie absurd. Two of the models concurred, but the GPT-4o model seemed to simply take the comment at face value, reasoning that "The commenter is suggesting that anyone who advocates for stopping climate change is a hypocrite". This points to the complexity of understanding nuanced concepts such as irony or sarcasm. 
\end{enumerate}

\end{document}